\documentclass[runningheads]{llncs}
\usepackage[T1]{fontenc}

\usepackage{graphicx}
\usepackage{xcolor}
\usepackage{booktabs}
\usepackage{multicol} 
\usepackage{amsmath}
\usepackage{hyperref}

\usepackage{adjustbox} 
\begin{document}
\title{Visual Quality Score Assessment of Large White Goods in Remanufacture with Multi-View Deformable-DETR }
\titlerunning{Visual Quality Score Assessment of Large White Goods in Remanufacture}
%
\author{Paul Koch\inst{1}\orcidID{0000-0002-7045-5700} \and
Vivek Chavan\inst{1}\orcidID{0000-0001-9350-5259}}
\authorrunning{Koch and Chavan}
%
\institute{Fraunhofer-Institut für Produktionsanlagen und Konstruktionstechnik (IPK), Pascalstraße 8-9, 10587 Berlin, Germany \\
\textbf{Email:} \email{paul.koch@ipk.fraunhofer.de}\\
\textbf{Website:} \url{https://www.ipk.fraunhofer.de/}\\
\textbf{Code:} \url{https://github.com/KochPJ/SSLMV}}
\maketitle              
\begin{abstract}

Remanufacturing large white goods is essential for a circular economy, yet visual quality assessment remains a manual bottleneck for training and pricing. Conventional detection methods require extensive annotation and struggle with small defects in high-resolution multi-view data. We present a multi-view framework based on Deformable-DETR for automated quality scoring that aggregates information across redundant views to extract fine-grained features. To enhance robustness with limited labels, we employ self-supervised pretraining followed by supervised fine-tuning on expert-annotated scores. Additionally, a linear projection over frozen feature maps identifies regions of interest to explain model decisions. Evaluated on an industrial multi-view dataset, our approach delivers precise quality assessments while reducing reliance on manual annotation and per-part customization, enabling scalable and transparent inspection for remanufacturing lines.
\end{abstract}
\section{Introduction}
The white goods industry faces significant resource inefficiency and increasing regulatory pressure to shift from linear to circular models. New frameworks, such as the EU’s Ecodesign for Sustainable Products Regulation (ESPR) and Digital Product Passport (DPP), mandate a more sustainable lifecycle for large appliances. Among other, a primary bottleneck in this transition is visual quality assessment, which determines if a product is suitable for refurbishment, resale, or recycling. Current manual inspection is time-consuming, subjective, and difficult to scale, particularly when evaluating appliances from multiple viewpoints. Automated, reliable quality scoring is therefore essential to increase remanufacturing throughput.

This work is based on the KIKERP project (see acknowledgment), which addresses these challenges by developing a multimodal, AI-based system for evaluating end-of-life appliances. This system captures multi-view visual data and metadata to identify products and assign standardized quality scores. These scores support consistent decision-making for refurbishment and pricing, enabling data-driven remanufacturing workflows.

In this work, we present a multi-view architecture based on Deformable-DETR~\cite{deformdetr} (\underline{DE}tection \underline{TR}ansformer) that fuses view-based features into a unified quality scoring system. Our approach employs precise feature extraction with dense feature maps to enable heat map visualizations of the decision-making process via segmentation. We validate this method using a new multi-view quality scoring dataset collected within the KIKERP project featuring large white goods in manufacturing.

\section{Related Work}
\textbf{Vision in Remanufacturing.} Quality assessment is central to circular economy research. In automotive remanufacturing, Kaiser et al. \cite{corequality} proposed using robots for multi-perspective anomaly detection to assess core quality. However, anomaly detection often fails to distinguish between cosmetic variations and functional defects, making it difficult to scale to the visual diversity of consumer white goods. Schlüter et al. \cite{corecls} utilized CNN-based recognition with incremental learning to improve old core sorting. While these systems provide semi-automated support, they lack the fine-grained scoring necessary for complex triage and pricing.

\textbf{Visual Quality Assessment.} Industrial inspection primarily employs deep learning for defect localization and segmentation \cite{detr}. These methods require extensive annotation and yield low-level outputs that do not directly facilitate high-level pricing decisions. Unsupervised anomaly detection mitigates annotation costs by identifying deviations from "normal" distributions but typically produces binary labels rather than graded quality scores. Current research either aggregates discrete defect attributes \cite{qual_assess} or applies global perceptual scoring. However, most existing methods rely on single-view inputs and focus on detection rather than integrated multi-view scoring \cite{bauer2025dataset,nwankpa2021remanufacturing}, leaving a gap in scalable, transparent assessment.

\textbf{Multi-View (MV) Fusion.} Fusing perspectives for common prediction is prominent in 3D classification. While fusion can occur at various stages \cite{mvcnn}, recent models favor late-fusion of high-level features in the latent space \cite{mvip}. However, Zhu et al. \cite{deformdetr} argue that late-fusion can lose pixel-level accuracy. Their Deformable-DETR architecture utilizes a multi-scale deformable attention mechanism to query regions of interest across hierarchical feature maps, providing higher precision for small features. Training such architectures requires specialized datasets; however, most existing MV datasets focus on 3D reconstruction or classification rather than quality scoring.

\textbf{SSL and Vision Transformers (ViT).} While CNNs perform well in industrial tasks ViTs \cite{vits} offer a more powerful encoder architecture but require massive datasets and compute to generalize. Self-supervised learning (SSL) provides a solution; pretrained encoders like DINOv2 \cite{dinov2} produce robust, generalized features that perform well in production environments with minimal fine-tuning. By leveraging frozen SSL encoders, models can maintain high-level feature extraction while adapting to specific downstream tasks with limited labeled data.

\section{Methods}
This section details the Multi-View (MV) Deformable-DETR architecture and training protocol for visual quality assessment. We also describe a linear projection method that uses extracted feature maps to generate heatmaps, localizing quality-relevant features (e.g., surface defects). These visualizations provide explainability, enabling operators to audit model decisions. Finally, we present our MV dataset and experimental results.

\subsection{MV Deformable-DETR Architecture and Training}

\begin{figure}
    \centering
    \includegraphics[width=1.0\linewidth]{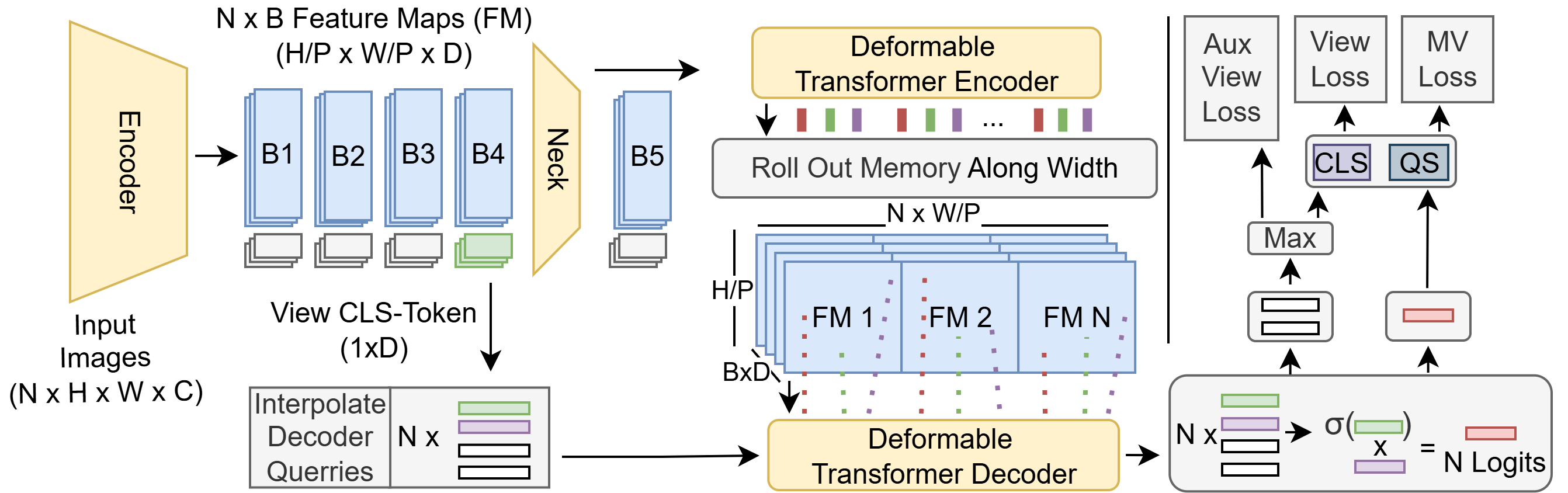}
    \caption{\textbf{MV Deformable DeTR}: Illustration of the architecture. }
    \label{fig:mvdeformarch}
\end{figure}

Standard DETR models~\cite{detr,deformdetr} target single-image tasks using four primary modules: an image encoder, a transformer encoder, a transformer decoder, and a task head. We adopt this framework but integrate a supplementary \textit{neck} module into the image encoder (see Fig.~\ref{fig:mvdeformarch}).

This neck adds encoder layers atop the existing feature extractor. By freezing the pretrained backbone (e.g., DINOv2) and fine-tuning only the neck, we leverage robust, generalized representations without altering the original weights. Post-training, the neck enables task-specific dense predictions, such as defect segmentation. This module is optional and can be omitted if the backbone is trained end-to-end or if high-resolution feature maps are unnecessary.

The MV adaptation of Deformable-DETR introduces two primary modifications: (A) spatial expansion of feature map embeddings for the transformer encoder-decoder, and (B) adaptive interpolation of decoder queries. In (A), we extend the hierarchical tokens extracted by the transformer encoder along the sequence dimension to accommodate $N$ input images. For (B), we interpolate fixed decoding queries to match the $N$ views and concatenate view-specific [CLS] (classification) tokens with $N \times A$ auxiliary tokens ($A$ being auxiliary tokens per view). These auxiliary tokens act as query registers~\cite{registers} to locate relevant features within the deformable memory. While more auxiliary tokens improve fine-grained feature extraction, they increase computational complexity.

Each [CLS] token encodes high-level view content, assisting the decoder in weighing perspective relevance. Simultaneously, decoding queries function as learnable latents that guide feature extraction alongside the auxiliary tokens. For final fusion, we apply a sigmoid activation to the [CLS] tokens and compute their weighted product with the corresponding queries. A max-pooling operation across the view dimension, followed by a fully connected layer, then outputs either MV classifications or a continuous visual quality score. Additionally, individual output tokens can generate view-specific assessments.

\begin{figure}
    \centering
    \includegraphics[width=0.8\linewidth]{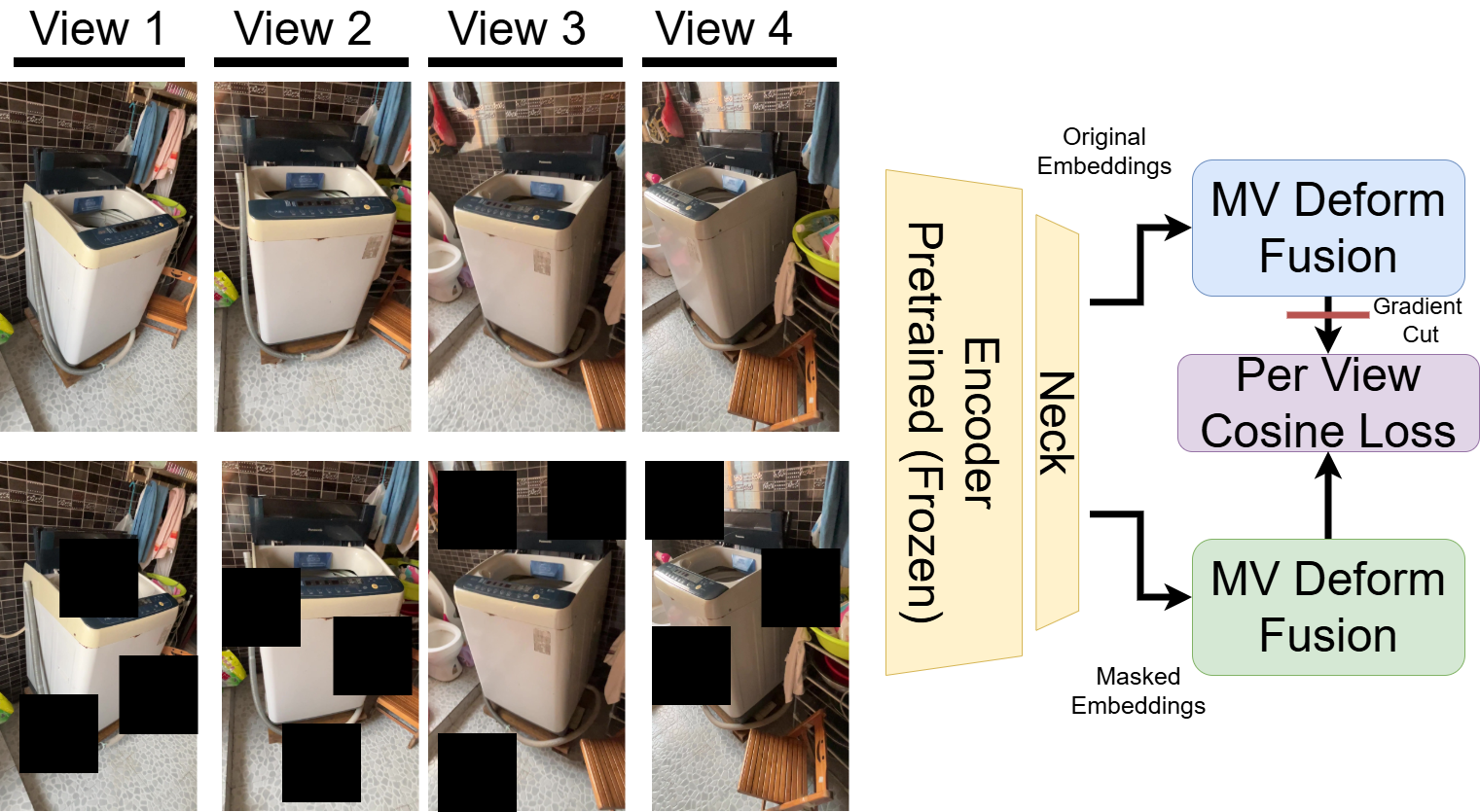}
    \caption{\textbf{SSL pretraining}: We use DinoV2 to extract the per view cls-tokens. Afterwards we have the MV-Deform architecture usining MiM-based SSL~\cite{mim} to reconstruct the cls-tokens based on masked inputs.}
    \label{fig:ssltrain}
\end{figure}

\textbf{SSL Pretraining:} As illustrated in Fig.~\ref{fig:ssltrain}, we leverage these view-wise outputs during self-supervised learning (SSL) pretraining to adapt the architecture. This allows the model to extract robust MV features from initial encoder embeddings using the large-scale MVImgNet dataset~\cite{MVimgnet}.

\begin{figure}
    \centering
    \includegraphics[width=1.0\linewidth]{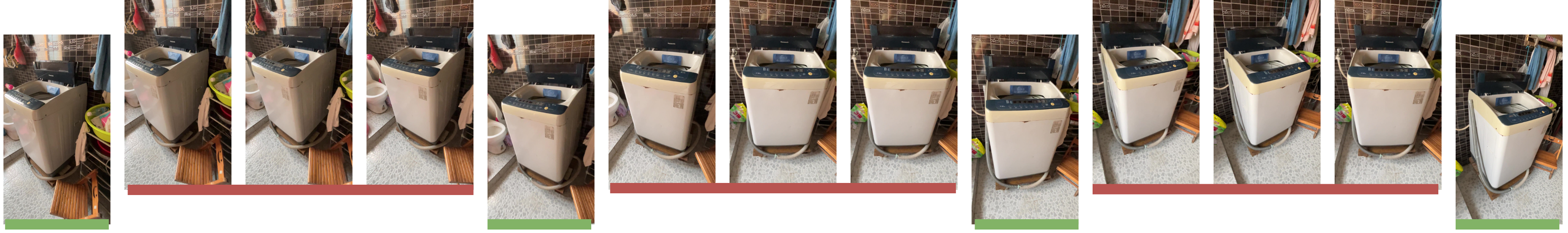}
    \caption{Data curation and keyframe selection based of 180° sample from MVimgnet~\cite{MVimgnet} based on methods by \cite{adcm}.}
    \label{fig:frameselection}
\end{figure}

MVImgNet comprises approximately $220\text{k}$ videos of unique objects across $238$ categories, captured from viewpoints spanning at least $180^{\circ}$. We subsample these dense arrays via keyframe selection (see Fig.~\ref{fig:frameselection}) to input exactly $8$ representative frames per scene. Temporal sampling jitter around these keyframes enhances training diversity. Pretraining utilizes a batch size of $32 \times 8$ images at a $640 \times 480$ resolution, a learning rate of $10^{-5}$, and a cosine similarity reconstruction loss.

\begin{figure}
    \centering
    \includegraphics[width=1.0\linewidth, trim={0 0 0 0cm}, clip]{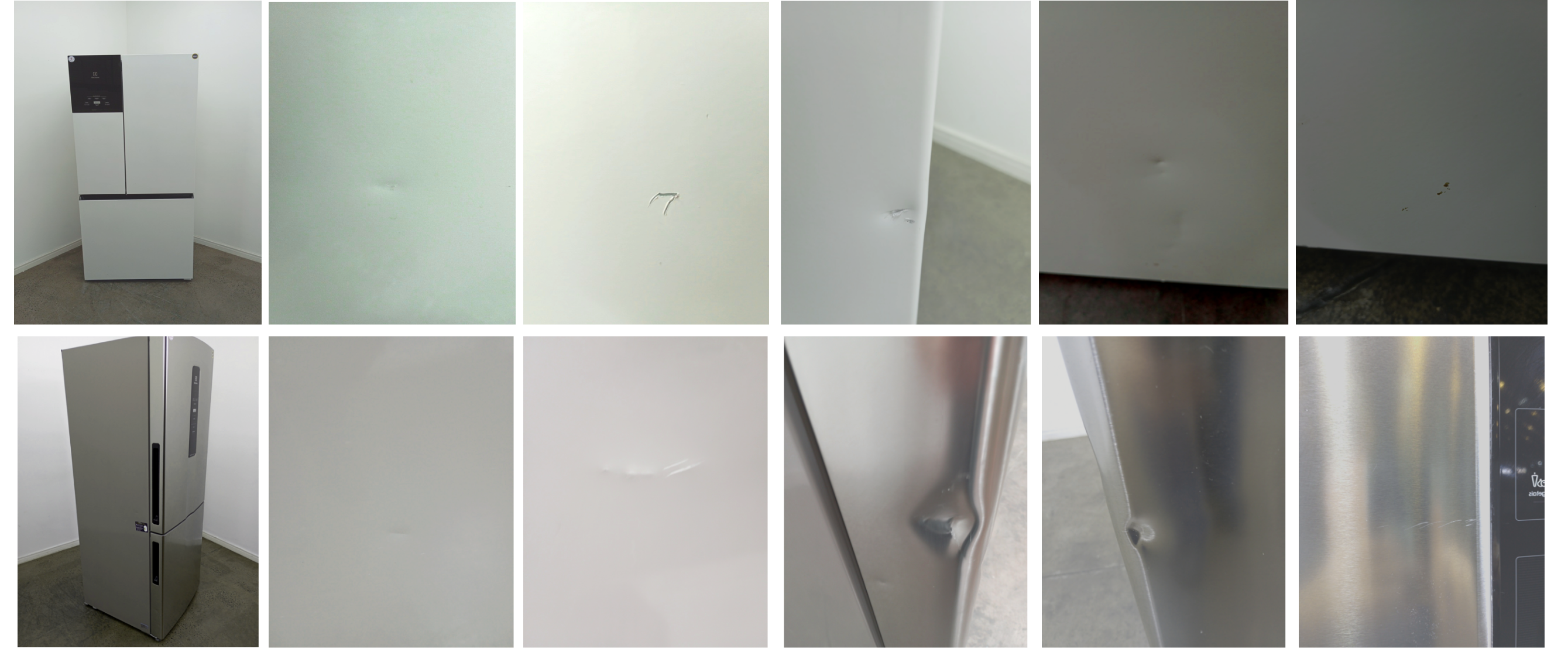}
    \caption{Sample devices (left) and close-up views of the visual defects.}
    \label{fig:kikerpexample}
\end{figure}

\textbf{KIKERP Dataset:} As no public dataset exists for KIKERP's quality scoring framework, we collected a novel MV dataset on-site at a Brazilian remanufacturing facility. The dataset (see Fig.~\ref{fig:kikerpexample}) features individual stock keeping units (SKUs) captured in a photobooth.

Each unit is photographed using a mobile phone from the left, center, and right perspectives, including a total-view image capturing the full profile of the appliance. Additionally, close-ups of all defects are recorded alongside final quality scores annotated by expert operators. We collected $23.3\text{k}$ images ($1920 \times 1440$ resolution) from $2,298$ SKUs, documenting $16.5\text{k}$ unique damage instances (ranging from zero to $30$ defects per unit). The data spans $406$ SKU categories across eight appliance types: stoves, refrigerators, freezers, washing machines, dishwashers, water dispensers, microwaves, and micro-ovens.

The SKUs are categorized into three quality groups: Very Good ($1,015$), Good ($907$), and Okay ($376$); unacceptable units are discarded upstream. Training follows the SSL methodology but optimizes a supervised Mean Squared Error (MSE) loss. This formulation treats the task as ordinal regression, preserving inherent inter-class distances. Discrete scores are converted to continuous targets via $\frac{n}{N+1}$ ($n$: class index, $N$: total classes). For $N=3$, targets align at $25\%$ intervals, enabling the model to represent uncertainty between adjacent discrete classes. We utilize a $90\%/10\%$ train/test split, reserving $230$ unseen SKUs for evaluation.

\begin{figure}
    \centering
    \includegraphics[width=1.0\linewidth]{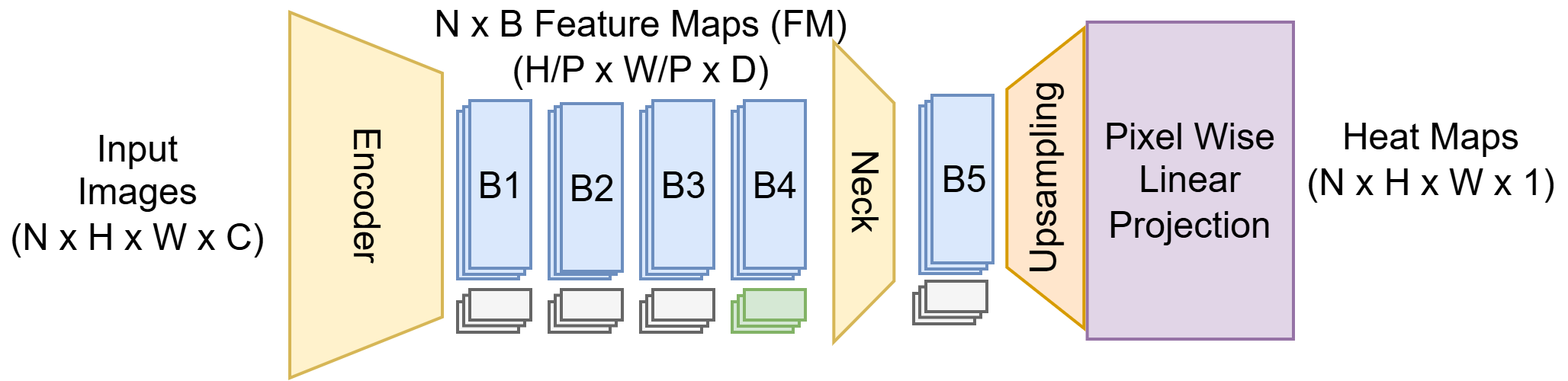}
    \caption{\textbf{Segmentation Architecture}: Using a single linear projection to extract heat maps highlighting areas of interest based on the frozen genearlized DinoV2 encoder with task specific neck extension.}
    \label{fig:segmentation}
\end{figure}

\begin{figure}
    \centering
    \includegraphics[width=1.0\linewidth]{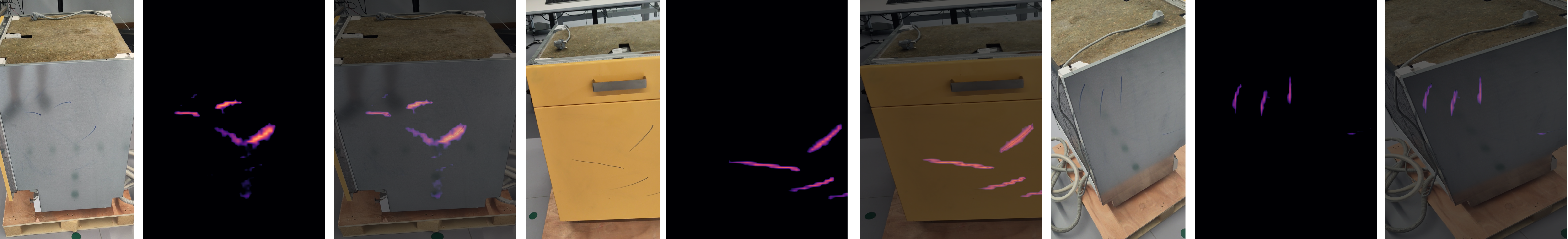}
    \caption{\textbf{Segmentation Illustration:} the linear projection can be trained fast and efficiently on few segmentation annotations to produce precise heat maps to pinpoint areas of interest to explain the quality scoring decision making.}
    \label{fig:segims}
\end{figure}

\textbf{Segmentation:} We manually annotated $590$ randomly selected images from the training set with pixel-wise segmentation masks to identify regions containing defects or other quality-relevant features. To localize the features driving the model's predictions, we trained a single fully connected linear projection layer to map the high-dimensional feature maps from the encoder neck to these segmented regions (see Fig.~\ref{fig:segmentation}). This task utilized an $80/20$ training-to-testing data split. To align the ground-truth labels with the spatial resolution of the feature maps, the segmentation masks were downsampled using bilinear interpolation to generate continuous (non-binary) heatmaps for supervised regression. During inference, these predicted maps are upsampled to their original dimensions to identify the pixel-wise areas of interest that justify the final quality score (see Fig.~\ref{fig:segims}). 

\textbf{Results:} Since our visual quality scoring (QS) approach is novel and not yet established in the literature, we evaluate our framework against state-of-the-art (SOTA), view-independent multi-view (MV) classification architectures~\cite{mvcnn,mvip}. These baseline methods feature similar architectural designs and can be easily adapted to output a continuous MV quality score rather than a categorical classification. We report the regression accuracy using the Mean Absolute Error (MAE), including a normalization relative to the quality class bin size ($25\%$) to indicate the average distance of predictions from the quality class center. Furthermore, we provide classification and segmentation accuracy metrics for a comprehensive evaluation. In Table~\ref{tab:quality_scoring}, we summarize the results of an ablation study alongside comparative analyses against SOTA implementations. For structural context, Table~\ref{tab:modelparams} details the parameter counts of our individual modules relative to a SOTA Transformer (TF) head approach~\cite{mvip}.

\begin{table}[h]
    \centering    
    \caption{\textbf{Ablation Study}: Comparing MV-Deform DeTR vs. SOTA and architecture modules. Best results in \textbf{Bold}.}
    \label{tab:quality_scoring}
  \begin{adjustbox}{width=\textwidth, center}
    \begin{tabular}{l || c | c || c | c | c | c || c | c | c | c | c }
        &\multicolumn{2}{c||}{Regression}&\multicolumn{4}{c||}{Classification} & \multicolumn{5}{c}{Segmentation}\\
         Method&  MAE & $\frac{\text{MAE}}{p=25\%}$& Top1 & P & R & F1 & mIoU & Top1 & P & R & F1 \\
         \toprule
         CNN\cite{mvcnn} & 10.32 & 41.28 &65.0 & 52.08 & 48.15 & 44.51 & 48.05 & 96.11 & 48.05& 50.00 & 49.01 \\ 
         \midrule
         DinoV2~\cite{dinov2} &8.54 & 34.16 & 65.0 & 44.44 & 48.15 & 45.48 & -& -& -& -& -\\
         ~~~~ +Neck (Ours) & 8.87& 35.48 & 65.0 & 46.81 & 48.15 & 44.80 & 62.27 & 96.69 & 80.99 & 66.09 & 70.95 \\ 
         ~~~~ +TF~\cite{dinov2} &7.94& 31.76 &67.50& 80.86& 56.48 & 59.26& 61.83 & 96.64 & 80.5 & 65.63 & 70.44\\ 
         ~~~~ +SSL (Ours) & 8.31 & 33.24 &70.0 & 80.0& 58.33 & 61.25 & 61.2 & 96.59 & 79.92 & 64.92 & 69.67 \\
         \midrule
         MV Deform (Ours) & \textbf{6.88} & \textbf{27.52} & 70.0& 80.01 & 58.33 & 61.25 & 62.08 & 96.63 &78.75& 66.44 & 70.75\\
         ~~~~+SSL (Ours) & 7.61 & 30.44 & \textbf{75.0} & \textbf{84.44} & \textbf{62.04} &\textbf{ 64.99} & \textbf{63.01} & \textbf{96.75}& \textbf{81.59} & \textbf{66.92} & \textbf{71.83}\\
         \bottomrule
    \end{tabular}
    \end{adjustbox}
\end{table}

\begin{table}[h]
    \centering
    \caption{\textbf{Model Parameter Counter}: Comparing our Deform-MV vs. a normal Transformer~\cite{mvip} used for view token fusion.}
    \label{tab:modelparams}
    \begin{tabular}{ c | c | c }
         Module & \multicolumn{2}{c}{Parameters}\\
         \toprule
         Vit-14-B~\cite{dinov2}& 86.6 Mio & 49.24\%\\
         Neck (Ours) & 14.18 Mio & 8.64 \% \\
         CLS \& QS & 5.91 Mio & 3.36\% \\
         \midrule
         MV-Deform (Ours) &69.16 & 39.34\% \\
         \multicolumn{3}{c}{vs.}\\
         Transformer~\cite{mvip} & 74.75 & 41.20\% \\
         \bottomrule
    \end{tabular}
\end{table}

\section{Conclusion}
Our results in Table ~\ref{tab:quality_scoring} show the increasing effect of our novel architecture modules, SSL pretraining, and MV-Deform DeTR architecture. However, these evidence are based on a rather small sized dataset with some inconsistencies in the quality scoring due to human error and disagreements in the scoring. We believe that these disagreement issues can be statistically removed by scaling up the dataset, but in the current form we find that the model is forced to overfit when trained to long, which is expected based on our inconsistency observations. However, with our results, we show a promising proof-of-concept of how to efficiently pretrain a MV architecture and fine-tune it to produce quality scores and explanatory heat maps. This approach is universal and can be adapted to any number of objects, views, object size, resolution, or visual defects.

\begin{credits}

\subsubsection{\ackname} This work is supported by the German Federal Ministry of Research, Technology and Space (BMFTR) and the German Aerospace Center (DLR) under the KIKERP project (Grant No.\ 16IS23055C) within the KI4KMU program.

\subsubsection{\discintname}
The authors have no competing interests to declare that are
relevant to the content of this article.
\end{credits}
%
%
%
%

\bibliographystyle{splncs04}
\bibliography{references}

\end{document}